\documentclass[conference]{IEEEtran}
\IEEEoverridecommandlockouts
\usepackage{cite}
\usepackage{amsmath,amssymb,amsfonts}
\usepackage{algorithmic}
\usepackage{graphicx}
\usepackage{subfigure}
\usepackage{float}
\usepackage{textcomp}
\usepackage{xcolor}
\usepackage{multirow}
\def\BibTeX{{\rm B\kern-.05em{\sc i\kern-.025em b}\kern-.08em
    T\kern-.1667em\lower.7ex\hbox{E}\kern-.125emX}}

\begin{document}

\title{
Identity-aware Feature Decoupling Learning for Clothing-change Person Re-identification}

\author{\IEEEauthorblockN{1\textsuperscript{st} Haoxuan Xu}
\IEEEauthorblockA{\textit{School of Artificial Intelligence} \\
\textit{Beihang University}\\
Beijing, China \\
xhaoxuan@buaa.edu.cn}
\and
\IEEEauthorblockN{2\textsuperscript{nd} Bo Li}
\IEEEauthorblockA{\textit{School of Artificial Intelligence} \\
\textit{Beihang University}\\
Beijing, China \\
boli@buaa.edu.cn}
\and
\IEEEauthorblockN{3\textsuperscript{rd} Guanglin Niu\(^*\)}
\IEEEauthorblockA{\textit{School of Artificial Intelligence} \\
\textit{Beihang University}\\
Beijing, China \\
beihangngl@buaa.edu.cn}
}

\maketitle

\begin{abstract}
Clothing-change person re-identification (CC Re-ID) has attracted increasing attention in recent years due to its application prospect. Most existing works struggle to adequately extract the ID-related information from the original RGB images. In this paper, we propose an Identity-aware Feature Decoupling (IFD) learning framework to mine identity-related features. Particularly, IFD exploits a dual stream architecture that consists of a main stream and an attention stream. The attention stream takes the clothing-masked images as inputs and derives the identity attention weights for effectively transferring the spatial knowledge to the main stream and highlighting the regions with abundant identity-related information. To eliminate the semantic gap between the inputs of two streams, we propose a clothing bias diminishing module specific to the main stream to regularize the features of clothing-relevant regions. Extensive experimental results demonstrate that our framework outperforms other baseline models on several widely-used CC Re-ID datasets.
\end{abstract}

\footnotemark
\footnotetext{* corresponding author. This work was supported by the National Natural Science Foundation of China (No. 62376016).}


\section{Introduction}
Person re-identification (Re-ID) aims at matching the same pedestrian across different cameras \cite{c16}. Most existing methods predominantly utilize global representations for matching, which are only applicable to pedestrians without clothing change \cite{c10, c11,c12,c18,c19,c20,c21,c22,c26,c27}. Whereas, a person changing clothing is a widespread phenomenon in practice. Consequently, the more challenging task clothing-change person re-identification (CC Re-ID) has received significant attention recently, which attempts to associate the same pedestrian with changed clothes.

\begin{figure}[h]
    \label{fig1}
    \centering
    \includegraphics[width=8cm]{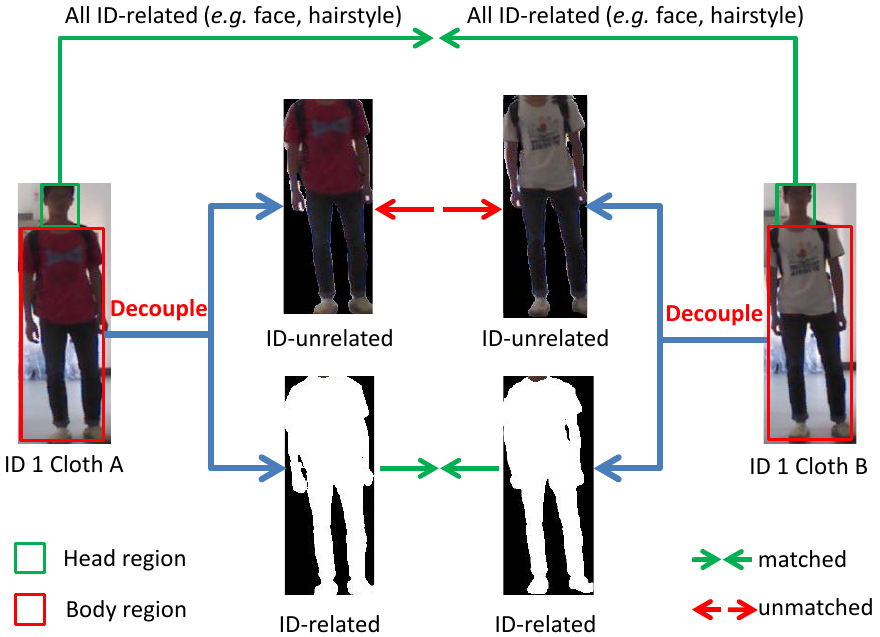}     
\caption{Illustration of the ID-related features distribution. The head regions contain purely ID-related features, while ID-related features and ID-unrelated features are coupled in the body regions.}

\end{figure}

\begin{figure*}[h]
    \centering
    \includegraphics[width=17cm]{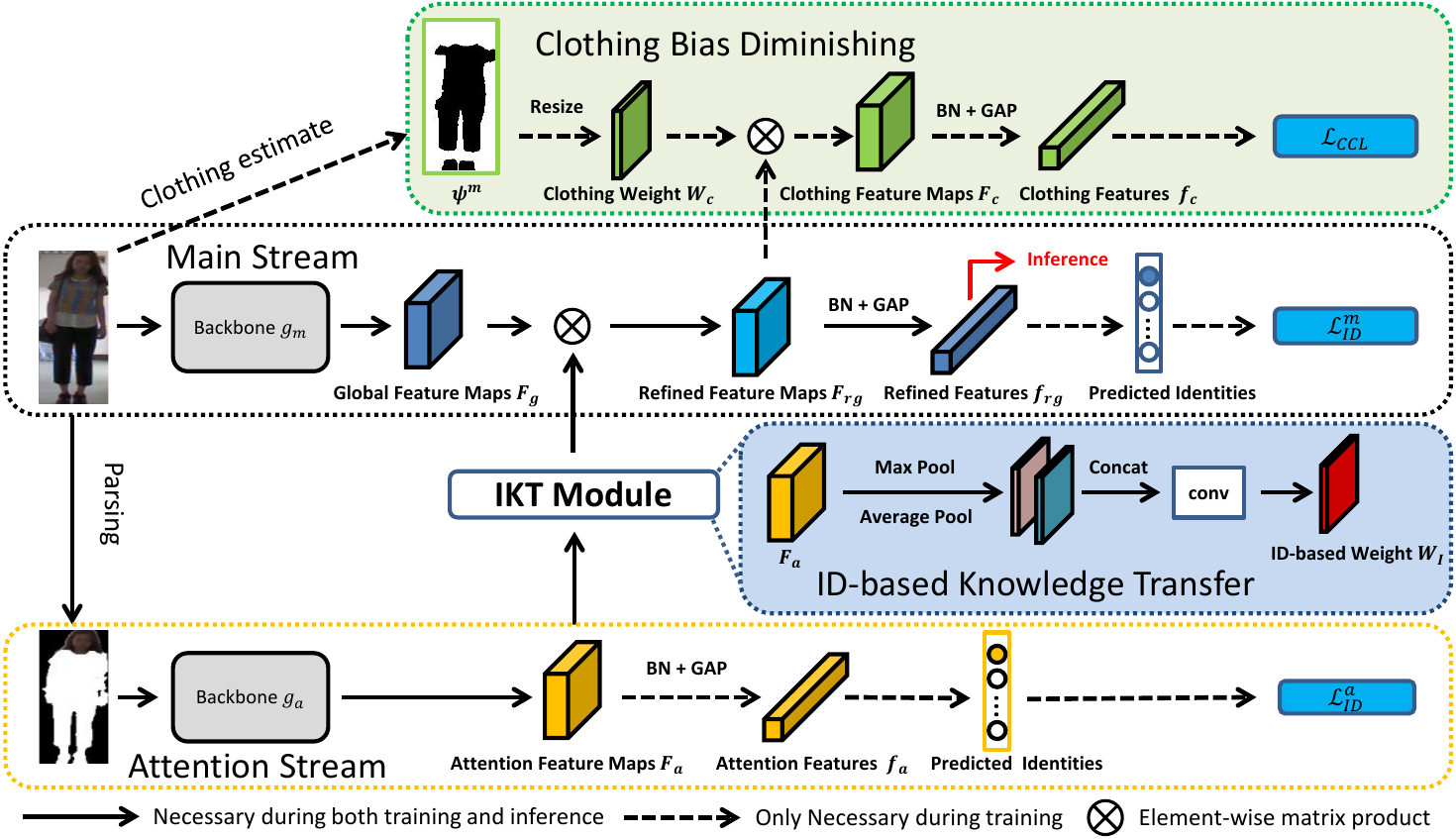}
\caption{Overview of our proposed IFD, which consists of an attention stream and a main stream. The attention stream learns a weight matrix with high values for identity-relevant regions and low values for identity-irrelevant regions at the feature level. The main stream aims to learn ID-related features under the guidance of the attention stream and clothing bias diminishing module.}
\label{fig2}
\end{figure*}


Driven by different motivations, existing methods for CC Re-ID can be broadly classified into two categories: \textbf{1)} data augmentation methods and \textbf{2)} biometrics-based methods. Data augmentation methods argue that the scale of current CC Re-ID datasets is insufficient to fully capture identity-related (ID-related) features and attempt to augment training data \cite{c3, c15, c23}. However, these methods are subject to the quality of the generated virtual data, and their effectiveness is typically difficult to interpret. On the other hand, biometrics-based methods aim to explicitly capture stable biometrics features. These methods can be further subdivided into multi-modality and single-modality methods. The multi-modality methods exploit extra modalities as auxiliary information to highlight ID-related features. Wang \textit{et al.} enhance ID-related features by extracting face features \cite{c17}. Methods such as SPT+ASE, CESD, FASM, and GI-ReID learn ID-related features utilizing sketches, keypoints, human contours, and gait information, respectively \cite{c4, c5, c8, c9}. Additionally, some studies argue that the 3D shape contains rich ID-related information and attempt to leverage 3D information \cite{c6, c13}. In contrast, single modality methods such as CAL, RCSANet and AIM attempt to directly mine ID-related features by clothing adversarial loss, clothing status awareness and causality, respectively \cite{c2, c7, c32}.

In the real world, humans can recognize their acquaintances through various identity clues (\textit{e.g.}, face, hairstyle, body shape, height, gait) even if these individuals wear unfamiliar clothing. However, certain clues, such as height and gait, cannot be reliably estimated from a single image, and the remaining clues are distributed across both clothing-irrelevant regions (\textit{i.e.} head regions) and clothing-relevant regions (\textit{i.e.} body regions). As illustrated in Fig. 1, the head regions primarily contain ID-related features, such as the face, hairstyle and head contour. In contrast, the body regions exhibit a mixture of identity-related and identity-unrelated features, which can lead to erroneous match results. Thus, the key challenges of CC Re-ID can be concluded: \textbf{1) leveraging only specific part of ID-related features can not comprehensively represent an individual, 2) clothing-relevant regions actually couple the implicit ID-related features, which is hard to be extracted for CC Re-ID.} Although existing approaches mitigate the effect of clothing changes to some extent, they still lack effective constraints to ensure that the model consistently focuses on all critical ID-related features. Specifically, each of the aforementioned multi-modality methods tends to capture only one category of ID-related features, leaving other critical discriminative ID-related features underutilized. As for single-modality methods, they can only extract coarse ID-related features by overlooking the semantics of body parts. 






To enhance the ID-related features for CC Re-ID, we propose a novel \underline{\textbf{I}}dentity-aware \underline{\textbf{F}}eature \underline{\textbf{D}}ecoupling (IFD) learning framework that consists of an attention stream and a main stream. Both streams employ the same backbone for feature extraction but operate independently without shared weights. The attention stream processes clothing-masked images, while the main stream takes the original images as input. To ensure that the main stream focuses comprehensively on the regions implying identity information, we incorporate an ID-based Knowledge Transfer (IKT) module between the two streams. Additionally, to decouple the ID-related features from the clothing-relevant region, we introduce a Clothing Bias Diminishing (CBD) module, which helps model the consistent clothing features with regard to the same individual. 


In summary, our contributions are listed as follows:

\begin{itemize}
    \item We are the first to propose a dual-stream identity-attention model that effectively compels the network to focus comprehensively on the regions containing distinctive identity information.
    \item An effective CBD module is developed to maintain the consistency of clothing features for the same individual.
    \item Extensive experiments demonstrate that our method achieves state-of-the-art on several clothing-change Re-ID datasets including PRCC and LTCC \cite{c8,c9}.
\end{itemize}

\section{Methodology}

IFD aims to comprehensively mine the ID-related features to address the CC Re-ID issue. As illustrated in Fig. 2, the framework begins by extracting a clothing-masked image \textit{M} from the input image \textit{I} using an off-the-shelf human parsing network \cite{c14}.  The clothing-masked image contains the critical head and contour information, which is essential for learning ID-related features, but it discards all the color information of body parts, which is critical to re-identify persons in the conventional Re-ID scenarios. To robustly extract ID-related features, we implement a dual-stream framework with an ID-based knowledge transfer module to guide the main stream toward comprehensively emphasizing ID-related regions.


The IKT module can help locate the ID-related regions. However, the semantic gap between the inputs of the two streams may introduce error. The body regions of inputs merely contain shape information in the attention stream, while clothing-relevant features can still couple with ID-related features in body parts of main stream. Thus, the IKT module might inadvertently amplify the influence of ID-unrelated features while enhancing ID-related features, which can limit the overall performance of our model. To guarantee that the final features are exclusive of clothing color and texture, we introduce a CBD module.

\subsection{ID-based Knowledge Transfer}
Motivated by the spatial attention mechanisms that are widely used in computer vision to capture fine-grained local features and precisely locate task-relevant regions, we attempt to enhance the attention toward ID-related regions for the CC Re-ID task\cite{c25, c24}. However, it is naive to directly utilize spatial attention due to the difficulty of learning effective spatial attention weights without auxiliary supervision. To address this, we propose an ID-based Knowledge Transfer module that facilitates the learning of robust and effective spatial attention weights.

As shown in Fig. 2, we design a mutual learning framework with two stream,  \(g_{m} \left ( . \right ) \) and \(g_{a} \left ( . \right ) \) denotes the backbone of main stream and attention stream, respectively. The original image \(I\) is passed through \(g_{m} \left ( . \right ) \) to extract the global feature maps \(F_{g}\). Simultaneously, the masked image \(M\) is fed into \(g_{a} \left ( . \right ) \) to obtain the attention feature maps \(F_{a}\). Taking \(F_{a}\) as input, our IKT module derives ID-based attention matrix \(W_{I}\): 
\begin{equation}
W_{I} = \sigma\left ( W_{conv}\ast \left [ mp\left ( F_{a} \right ); ap\left ( F_{a} \right )\right ] \right)
\end{equation}
where \(mp\) denotes max pooling along the channel, \(ap\) denotes average pooling along the channel, \(\ast\) denotes convolution operation, \(W_{conv}\) indicates the weights of convolution filters, and \(\sigma\left ( . \right ) \) denotes the sigmoid function. The ID-based attention matrix is then applied to the global feature \(F_{g}\) as formulated:
\begin{equation}
F_{rg} = W_{I} \otimes F_{g}
\end{equation}
where \(\otimes\) denotes the element-wise matrix product, \(F_{rg}\) denotes refined feature maps. 

\subsection{Clothing Bias Diminishing Module}

To decouple the ID-related features in clothing-relevant regions, we propose a Clothing Bias Diminishing module. As illustrated in the top of Fig. 2, we estimate a fine-grained mask \(\psi ^{m}\) for clothing-relevant parts from the original image \(I\). The pixel value of \(\psi ^{m}\) can be formulated as:
\begin{equation}
\psi_{\left ( i,j \right ) }^{m} = \begin{cases}
 1 & \text{,  if  } I_{i,j} \in C \\
 0 & \text{,  Otherwise } 
\end{cases}
\end{equation}
where \(C\) denotes the set of clothing part categories. Then we resize the \(\psi ^{m}\) to match the dimensions of \(W_{I}\), resulting in \(W_{c}\), which is used to perform an element-wise matrix product with \(F_{rg}\) to derive the clothing-related feature maps \(F_{c}\). By applying batch normalization and global average pooling operation to \(F_{c}\), we obtain the clothing features \(f_{c}\).

To restrain the contribution of clothing at the feature level, we propose Clothing Contrastive Loss \(\mathcal{L}_{CCL}\). Let \(i\) be the index of an arbitrary sample in a batch, the \(\mathcal{L}_{CCL}\) is defined:


\begin{equation}
\mathcal{L}_{CCL}  = -\frac{1}{N} \sum_{i = 1}^{N}\frac{1}{\left | P_{i} \right | } \sum_{p\in P_{i}}w_{p}\log\frac{e^ {\left ( f_{c}^i\cdot f_{c}^p /\tau \right ) }}{e^ {\left ( f_{c}^i\cdot f_{c}^p /\tau \right ) }+\sum\limits_{j\in N_i}e^ {\left ( f_{c}^i\cdot f_{c}^j /\tau \right ) } }  
\end{equation}
\begin{equation}
w_p = \begin{cases}
\frac{1}{T}   & \text{ ,  if } c_{p}\ne c_{i} \\
1  & \text{ ,  Otherwise } 
\end{cases}
\end{equation}
where \(N\) is the batch size,  \(P_{i} \left( N_i \right)\) denotes the set of samples with the same ID as (different ID from) \(i\), and \(\tau\) is a temperature parameter. \(T \in R^{+} \) is a variable parameter, and \(c_{x}\) denotes the clothing label. Equation (5) serves as an incentive function, encouraging the network to focus more on pairs with the same ID label but different clothing labels during training.


\begin{table*}[]
\centering
\caption{Comparison Results with state-of-the-art methods on the PRCC and LTCC dataset(\%), where 'sketch', 'sil', '3D', 'pose', 'parsing','aug' denote the contour sketches, silhouettes, 3D shape, keypoints, human parsing and data augmentation, respectively. The best results are indicated by \textcolor{red}{red}.}
\label{table1}
\begin{tabular}{l|c|cccc|cccc}
\hline
\multicolumn{1}{r|}{\multirow{3}{2.5cm}{\centering Method}} & \multirow{3}{2.5cm}{\centering Modality} & \multicolumn{4}{c|}{PRCC}                                                                                      & \multicolumn{4}{c}{LTCC}                                                                                      \\ \cline{3-10} 
\multicolumn{1}{r|}{}                         &                           & \multicolumn{2}{c|}{SC Mode}                                & \multicolumn{2}{c|}{CC Mode}                               & \multicolumn{2}{c|}{General Mode}                                & \multicolumn{2}{c}{CC Mode}                               \\ \cline{3-10} 
\multicolumn{1}{r|}{}                         &                           & \multicolumn{1}{l}{rank-1} & \multicolumn{1}{l|}{mAP}  & \multicolumn{1}{l}{rank-1} & \multicolumn{1}{l|}{mAP} & \multicolumn{1}{l}{rank-1} & \multicolumn{1}{l|}{mAP}  & \multicolumn{1}{l}{rank-1} & \multicolumn{1}{l}{mAP} \\ \hline
PCB \cite{c12}                                           & RGB                       & 99.8                       & \multicolumn{1}{c|}{97.0} & 41.8                       & 38.7                     & 65.1                       & \multicolumn{1}{c|}{30.6} & 23.5                       & 10.0                         \\
IANet \cite{c10}                                         & RGB                       & 99.4                       & \multicolumn{1}{c|}{98.3} & 46.3                       & 45.9                     & 63.7                       & \multicolumn{1}{c|}{31}   & 25.0                       & 12.6                    \\ \hline
SPT+ASE \cite{c9}                                       & sketch                    & 64.2                       & \multicolumn{1}{c|}{-}    & 34.4                       & -                        & -                          & \multicolumn{1}{c|}{-}    & -                          & -                       \\
CESD \cite{c8}                                          & RGB+pose                  & -                          & \multicolumn{1}{c|}{-}    & -                          & -                        & 71.4                       & \multicolumn{1}{c|}{34.3} & 26.2                       & 12.4                    \\
RCSANet \cite{c7}                                       & RGB                       & \textcolor{red}{100}                        & \multicolumn{1}{c|}{97.2} & 50.2                       & 48.6                     & -                          & \multicolumn{1}{c|}{-}    & -                          & -                       \\
AFD-Net \cite{c15}                                 & RGB+aug                       & 95.7                        & \multicolumn{1}{c|}{-} & 42.8                      & -                     & -                          & \multicolumn{1}{c|}{-}    & -                          & -                       \\
FASM \cite{c4}                                          & RGB+pos+sil               & 98.8                       & \multicolumn{1}{c|}{-}    & 54.5                       & -                        & 73.2                       & \multicolumn{1}{c|}{40.8} & 38.5                       & 16.2                    \\
GI-ReID \cite{c5}                                       & RGB+sil                   & 80.0                       & \multicolumn{1}{c|}{-}    & 33.3                       & -                        & 63.2                       & \multicolumn{1}{c|}{29.4} & 23.7                       & 10.4                    \\

CAL \cite{c2}                                           & RGB                       & \textcolor{red}{100}                       & \multicolumn{1}{c|}{99.8} & 55.2                       & 55.8                     & 74.2                       & \multicolumn{1}{c|}{40.8} & 40.1                       & 18.0                    \\
3DSL \cite{c6}                                          & RGB+3D                    & -                          & \multicolumn{1}{c|}{-}    & 51.3                       & -                        & -                          & \multicolumn{1}{c|}{-}    & 31.2                       & 14.8                    \\
3DInvarReID \cite{c13}                                          & RGB+3D                    & -                          & \multicolumn{1}{c|}{-}    & 51.6                       & 52.5                        & -                          & \multicolumn{1}{c|}{-}    & 37.8                      & 16.7 
           \\
AIM \cite{c32}                                         & RGB                       & \textcolor{red}{100}                       & \multicolumn{1}{c|}{\textcolor{red}{99.9}}    & 57.9                       & 58.3                        & 76.3                      & \multicolumn{1}{c|}{41.1} & 40.6                       & 19.1     
           \\
CCFA \cite{c3}                                          & RGB+aug                      & 99.6                       & \multicolumn{1}{c|}{98.7} & 61.2                      & 58.4                     & 75.8                       & \multicolumn{1}{c|}{42.5} & 45.3                       & 22.1                    \\ \hline
IFD                                    & RGB+parsing                   & 99.0                      & 
\multicolumn{1}{c|}{98.2} & \textcolor{red}{65.3}                      & \textcolor{red}{61.7}                   & \textcolor{red}{81.4}                      & \multicolumn{1}{c|}{\textcolor{red}{65.3}} & \textcolor{red}{64.3}                       & \textcolor{red}{42.3}                    \\ \hline
\end{tabular}
\end{table*}

Existing Re-ID methods typically employ PK sampling strategy during training, where \(K\) samples per ID and \(P\times K\) samples per batch, but PK sampling overlooks the appearance diversity. However, in this paper, we encourage the network to learn ID-related features through our proposed loss function \(\mathcal{L}_{CCL}\), and its effectiveness hinges on diversity in both persons and appearances within a batch. RAS sampling strategy can maintain appearance diversity by sampling \(A\) appearances of each person and fixed \(K\) images of each appearance in a batch \cite{c28}. whereas, this sampling strategy may discard a significant number of samples for appearances with large proportion by ignoring the maldistribution of different appearances. Thus, we modify RAS by replacing the fixed number of each appearance with a proportion number to accelerate the effect of \(\mathcal{L}_{CCL}\).


\subsection{Training}
We adopt ResNet-50 pre-trained on ImageNet as our backbone \cite{c1, c31}. During the training stage, we first train the attention stream with \(\mathcal{L}_{ID}^{a}\) to obtain effective feature maps with abundant ID-related information, and then we jointly train the dual streams under the guidance of the overall loss:
\begin{equation}
\mathcal{L}_{all} = \mathcal{L}_{ID}^{m} + \mathcal{L}_{ID}^{a} + \lambda \cdot \mathcal{L}_{CCL}
\end{equation}
where \(\lambda\) denotes the weight of the \(\mathcal{L}_{CCL}\), which is set to 1.0.

\section{Experiments}
\subsection{Datasets and Evaluation Protocols}
We mainly evaluate our proposed method on two popular CC Re-ID benchmark datasets PRCC and LTCC. We employ two frequently-used metrics rank-1 and mAP to perform the evaluation results. Three kinds of test settings are defined as following: (i) \textbf{general mode} (both clothing-change and clothing-consistent ground truth samples are used to evaluate accuracy), (ii) \textbf{same-clothing mode} (only clothing-consistent ground truth samples are used to evaluate accuracy), (iii) \textbf{clothing-change mode} (only clothing-change ground truth samples are used to evaluate accuracy). In terms of PRCC, we report the evaluation results of same-clothing mode and clothing-change mode. As for LTCC, the accuracies for general mode and clothing-change mode are provided.

\subsection{Comparison With State-of-the-art Methods}
We compare our IFD with two traditional Re-ID methods, namely PCB, IANet, and eleven clothing-change Re-ID methods, including SPT+ASE, CESD, RCSANet, AFD-Net, FASM, GI-ReID, CAL, 3DSL, 3DInvarReID, AIM, and CCFA. As is illustrated in Tab.~\ref{table1}, our proposed IFD achieves superior performance with 19.0\% / 4.1\% absolute improvements in rank-1 on LTCC/PRCC of CC Mode, illustrating the effectiveness of our method that effectively mines comprehensive ID-related information whereas other methods typically capture only a single category of ID-related information. In the SC mode of PRCC, the 99.0\% rank-1 of IFD is close to saturation but inferior to some methods. The reason is that our IFD aims to capture clothing-irrelevant features but there are only clothing-consistent ground truth samples in this mode.


\begin{table}[t]
\centering
\caption{The ablation studies of IFD on PRCC and LTCC.}
\label{table2}
\begin{tabular}{c|cc|cc}
\hline
\multirow{2}{*}{Method} & \multicolumn{2}{c|}{PRCC} & \multicolumn{2}{c}{LTCC} \\ \cline{2-5} 
                        & rank-1       & mAP        & rank-1       & mAP       \\ \hline
baseline       & 24.3         & 11.2       & 16.4         & 9.2       \\
w/ IKT                 & 50.3         & 49.2       & 50.7        & 24.6      \\
w/ CBD                  & 51.2         & 45.3       & 49.6         & 23.4      \\\hline
IFD w/ CL                    & 58.9         & 52.3       & 57.8         & 30.6      \\
IFD                     & 65.3         & 61.7       & 64.3         & 42.3      \\\hline
\end{tabular}
\end{table}

\subsection{Ablation Studies}
\textbf{The effectiveness of components.} To verify the effectiveness of each contribution in our framework IFD, we reproduce a baseline method (ResNet-50) that only remains \(\mathcal{L}_{ID}^{m}\) of IFD for training. As shown in Tab.~\ref{table2}, the two ablated models that integrate IKT or CBD to the baseline individually both improve the performance significantly compared with the baseline. Furthermore, the whole model IFD introducing both IKT and CBD into the baseline obtains superior performance than the two ablated models. These results illustrate that both IKT and CBD could effectively facilitate our model to decouple clothing-relevant features and enhance ID-related features.

\begin{figure}[t]
    \centering
    \includegraphics[width=9cm]{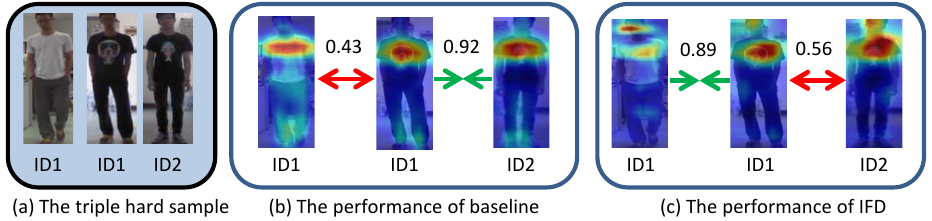}     
\caption{An intuitive comparison of the baseline and our model IFD specific to hard triples, namely the positive sample has absolutely different outfit with the anchor while the negative sample dressing similar with the anchor, together with their visualization results derived by grad-cam \cite{c30}.}
\end{figure}

\textbf{Comparison of CCL and standard contrastive loss.} We compare our clothing contrastive loss with widely used supervised contrastive loss \cite{c29}. As shown in Tab.~\ref{table2}, IFD with contrastive loss namely IFD w/CL is superior to the baseline. Besides, IFD with clothing contrastive loss namely IFD further outperforms IFD w/CL significantly, which illustrates the effectiveness of the designed clothing bias diminishing strategy.

\subsection{Visualization}
As the similarity between sample pairs and the heatmap visualization results shown in Fig. 3, the baseline believes the negative sample is more similar to the anchor since it focuses more attention on ID-unrelated features such as the texture of clothing. In contrast, our IFD can consistently highlight the critical ID-related head and human contour features, facilitating the correct matching results.

\section{conclusion}

In this paper, we propose a novel \underline{\textbf{I}}dentity-aware \underline{\textbf{F}}eature \underline{\textbf{D}}ecoupling learning framework for the CC Re-ID task. Our approach could effectively transfer the ID-based spatial knowledge into the main stream guided by the ID-related features derived from the attention stream and diminish the clothing bias to enhance the robustness and discriminate feature extraction capability under clothing variations. Extensive experiments demonstrate that our method achieves state-of-the-art performance on several widely used CC Re-ID datasets.

\newpage
\bibliographystyle{IEEEbib}
\bibliography{main}

\end{document}